\documentclass[journal]{IEEEtran}

\ifCLASSINFOpdf
   \usepackage[pdftex]{graphicx}
   \graphicspath{{Images/}{..}}
\else
   \usepackage[dvips]{graphicx}
   \graphicspath{{Images/}{..}}
\fi

\usepackage[cmex10]{amsmath}

\ifCLASSOPTIONcompsoc
  \usepackage[caption=false,font=normalsize,labelfont=sf,textfont=sf]{subfig}
\else
  \usepackage[caption=false,font=footnotesize]{subfig}
\fi

\usepackage{multirow}

\usepackage{url}
\usepackage{tikz}

\newcommand\copyrighttext{%
  \footnotesize This work has been submitted to the IEEE for possible publication. Copyright may be transferred without notice, after which this version may no longer be accessible.}
\newcommand\copyrightnotice{%
\begin{tikzpicture}[remember picture,overlay]
\node[anchor=south,yshift=10pt] at (current page.south) {\fbox{\parbox{\dimexpr\textwidth-\fboxsep-\fboxrule\relax}{\copyrighttext}}};
\end{tikzpicture}%
}

\begin{document}

\title{Combining geolocation and height estimation of objects from street level imagery}

\author{Matej~Ulicny,
        Vladimir~A.~Krylov,~\IEEEmembership{Senior Member,~IEEE},
        Julie~Connelly,
        and~Rozenn~Dahyot,~\IEEEmembership{Member,~IEEE}%
\thanks{This work was partly funded by the SFI Research Centres ADAPT (13/RC/2106\_P2) and IForm (16/RC/3872), and is co-funded by the European Regional Development Fund.}
\thanks{M. Ulicny and R. Dahyot are with the Hamilton Institute, Maynooth University, 
Ireland (e-mail: matej.ulicny@mu.ie; rozenn.dahyot@mu.ie).}%
\thanks{V.~A.~Krylov is with Dublin City University, School of Mathematical Sciences, DCU, Dublin 9, Ireland (e-mail: vladimir.krylov@dcu.ie).}%
\thanks{J.~Connelly is with ADAPT Centre, Trinity College Dublin, College Green, Dublin 2, Ireland (e-mail: julie.connelly@adaptcentre.ie).}}

\maketitle

\copyrightnotice

\begin{abstract}
We propose a pipeline for combined multi-class object geolocation and height estimation from street level RGB imagery, which is considered as a single available input data modality. Our solution is formulated via Markov Random Field optimization with deterministic output. The proposed technique uses image metadata along with coordinates of objects detected in the image plane as found by a custom-trained Convolutional Neural Network. Computing the object height using our methodology, in addition to object geolocation, has negligible effect on the overall computational cost. Accuracy is demonstrated experimentally for water drains and road signs on which we achieve average elevation estimation error lower than 20cm.
\end{abstract}

\begin{IEEEkeywords}
Height estimation, Street level imagery, Markov Random Field, Object geolocation.
\end{IEEEkeywords}

\section{Introduction}
\label{sec:intro}

\begin{figure*}[t]
  \centering
  \includegraphics[width=\linewidth]{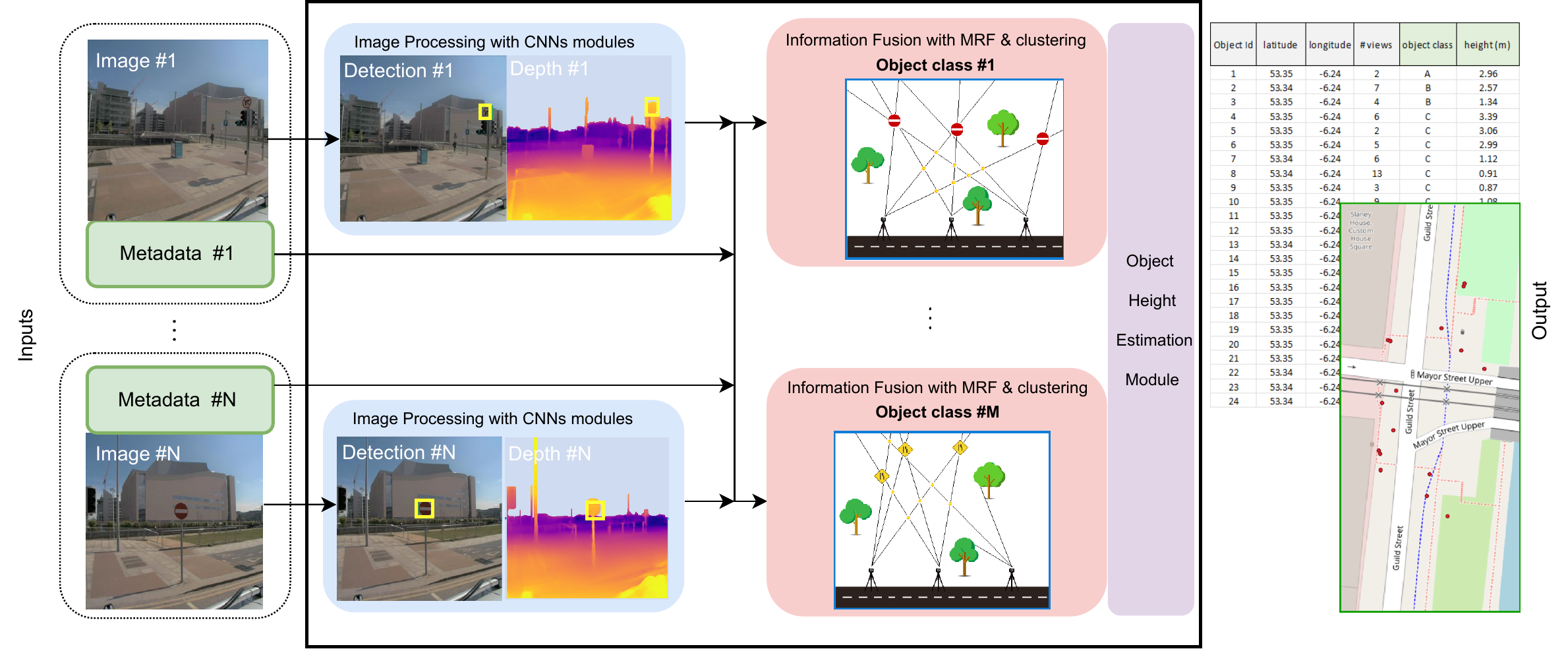}
  \caption{Our object geolocation pipeline augmented with multi-class object detection and object height estimation. Images are processed with  a CNN for object detection and a CNN for monocular depth estimation. Information fusion is performed  using triangulation aided by an MRF model and clustering step for each object class \cite{Krylov_2018}. For each geolocated object, its  height is estimated.}
  \label{fig:pipeline}
\end{figure*}

\IEEEPARstart{D}{etection}, recognition and geolocation of street furniture such as road signs and poles have historically gathered a lot of interest for management of the road network~\cite{Dahyot01,Timofte2009MultiviewTS} and driver assistance~\cite{6335478}.
Height estimation of these fixed elements in outdoor scenes is also of importance for instance  in the context of climate change events such as floods \cite{Ning_2022}. 
In this work we address object geolocation together with its height estimation.
Solutions for object geolocation and recognition have recently been  proposed using Convolutional Neural Networks (CNNs)  for image scene understanding, combined with graph based techniques for multi-sensor fusion such as camera GPS and IMU \cite{Krylov_2018,GeoGraphECCV2020}, (cf.  Sec. \ref{sec:SOTA}).
In Section \ref{sec:pipeline}, we  propose  a pipeline for object geolocation and height estimation using CNNs with Markov Random Fields (MRFs, cf. Fig. \ref{fig:pipeline}), with  Section \ref{sec:height} presenting the proposed new module for height estimation. Our approach is validated experimentally in Section \ref{sec:experiments} in a single-class object case (water drains, with custom made training image dataset), as well as in a multi-class object detection scenario (traffic signs). Our results show comparable or higher accuracy for geolocation as the state of the art approaches on a single class object  \cite{Krylov_2018,GeoGraphECCV2020} (about 1 meter precision) and a  competitive 20 centimeter accuracy for height estimation.

\section{Related Works}
\label{sec:SOTA}

\subsection{Object geolocation \& recognition in street level imagery}

Object detection and recognition have been performed using colour, edge and geometry information extracted from RGB images collected  from a camera mounted on a car rooftop~\cite{Dahyot01,5597423}. The GPS location of the car provides  a crude proxy estimate for the geolocation of the objects detected in the images. To complement detection in RGB  images, depth can also be  captured using LIDAR sensors or can alternatively be inferred from stereo-vision camera setups.  
More recently, Convolutional Neural Networks (CNNs) have been routinely used for both, object segmentation in images and for depth estimation from RGB images \cite{Krylov_2018,GeoGraphECCV2020}.
Krylov et al.~\cite{Krylov_2018} proposed the use of CNN segmentation to detect  telegraph poles and traffic lights in images using Google Street View (GSV) imagery. 
Their pipeline~\cite{Krylov_2018} also includes another pre-trained CNN for depth estimation~\cite{laina2016deeper}. 
Similarly Nassar et al.~\cite{GeoGraphECCV2020} uses an
ImageNet-pretrained CNN, to extract features that are then fed to both a classifier network and a regression network in order to predict the class of the objects of interest (trees and traffic signs) and regress the coordinates of its bounding box.

\subsection{Height estimation}
Ning et al.~\cite{Ning_2022} have utilised GSV images and their associated depth maps to infer door elevation that is important information  for building flood vulnerability assessment and insurance premium calculation for instance. 
Several studies~\cite{Diaz_2016,Yuan_2016,Zhao_2019,Al-Habashna_2021} have focused on estimating heights of buildings from street level images.  Diaz and Arguello~\cite{Diaz_2016} have used single view metrology, Yuan and Cheriyadat~\cite{Yuan_2016} have projected building footprints into images at ground elevation and found roofline elevation by measuring distances between local features. Zhao et al.~\cite{Zhao_2019} exploited building corner information for estimation of their height, while Al-Habashna~\cite{Al-Habashna_2021} also made use of building contours from Open Street Map data.
Wei et al.~\cite{isprs-archives-XLIII-B2-2021-557-2021} have used image plane to camera frame projection and homography estimation to predict heights of objects detected by a CNN from a single image.

\subsection{Graphs \& MRFs}
To merge the information extracted from multiple images capturing a scene,  Krylov et al.~\cite{Krylov_2018} defined a Markov Random Field on irregular grid with binary latent labels indicating occupancy of the site by an instance of the object class of interest  \cite{Krylov_2018}. Using triangulation, the sites are defined by intersecting rays (i.e. half-lines starting at the geolocated camera positions pointing in the direction of the detected objects identified in street level images) in the projected 2D planar world map. The MRF is optimised independently of the CNNs that are used in the pipeline to process input images.      

Recently, Graph Neural Networks \cite{GeoGraphECCV2020} have  been proposed for fusion of information from CNNs, providing an end-to-end learnable pipeline for geotagging static urban objects from multiple views. 
MRFs and graph formulations are related \cite{BookMRF2009,10.1007/978-3-319-46484-8_18}, and effectively solve the same problem of object re-identification in multiple views, where nodes correspond to detections in street level images and edges encode their visual and positional similarity across multiple images. The MRF setup allows for manual design of interaction terms (edges), and is thus capable of operating with little training data, unlike graph-NN-based designs, which require substantial amount of matched pairs of image segmentations with geolocations for each object type. Moreover, modular structure of CNN + MRF allows for independent optimization, and has more flexibility when new modules or constraints are integrated into the pipeline.
The explicit formulation of the MRF for geolocation~\cite{Krylov_2018} is therefore chosen in  this paper and extended  with the following innovations:\\[-12pt]
\begin{itemize}
\item image metadata and object coordinates in image plane are used to predict heights of the objects of interest (cf. Sec. \ref{sec:height}),
\item estimated heights are used for MRF optimization via a new unary energy term,
\item clustering process takes into account multiple object instances in close proximity (cf. Sec. \ref{sec:clustering}),
\item multiple classes of objects can be geolocated using a set of MRF models (cf. Sec. \ref{sec:MRF}), 
\item more recent and better performing CNN architectures than those used in~\cite{Krylov_2018} have been deployed (cf. Sec. \ref{sec:CNN}). 
\end{itemize}
Figure~\ref{fig:pipeline} summarizes our proposed pipeline with its additional  functionality for object height estimation.

\section{Pipeline} 
\label{sec:pipeline}
To estimate the height of an object, its GPS coordinates are first inferred and associated with multiple image views where that object has been detected~\cite{Krylov_2018}. Secondly, for each object, an estimate of its height can be computed for every image where it has been identified. The output of the processing pipeline is a list of objects detected along the considered roads, each described by its coordinates, height estimate and the number of images where this particular object has been identified. The height is reported as the mean (or median) height over single-view estimates along with its standard deviation (Fig. \ref{fig:pipeline}).

\subsection{Image processing with CNN modules}
\label{sec:CNN}
As an improvement to~\cite{Krylov_2018},  a cascade R-CNN model~\cite{cai2018cascade} with Res2Net-v1b-101 backbone~\cite{gao2019res2net}  for object detection locates objects of interest in images. This  CNN is trained for our objects of interest (water drains and traffic signs) and to tackle multi-class object detection (traffic signs). 
The second CNN predicts depth maps from images~\cite{Krylov_2018},  and the object segmentation masks are used to extract monocular distance from the depth maps. For monocular depth estimation we use AdaBins proposed in~\cite{Bhat_2021}.

\subsection{Information fusion with MRFs} 
\label{sec:MRF}
Object detections in images combined with the image metadata are used to construct rays (in 2D) that originate at the camera GPS coordinates (assumed identical to camera centre). The bearing (direction of ray) is selected towards the middle of the segmented object. 
The rays create a 2D graph with nodes formed by intersections of these rays. Edges are established between nodes on the intersected rays.
An MRF model is employed to estimate binary decisions ($z_i\in\lbrace 0,1 \rbrace$) indicating presence of the object of interest at a particular node $i$  \cite{Krylov_2018}.
The MRF energy of a node $z_i$ is governed by three unary terms and one pairwise term, hence considering all  $N_\mathcal{Z}$ nodes  of the graph:
\begin{equation}
\begin{split}
    \mathcal{U}\left(\mathbf{z}\right)= \sum_{i=1}^{N_\mathcal{Z}} [& \alpha \ u_0\left(z_i\right) + \beta \ u_1\left(z_i\right) 
     + \lambda \ u_2\left(z_i\right)] \\ & + \left(1-\alpha-\beta-\lambda\right) \sum_{i=1}^{N_\mathcal{Z}}\sum_{j=1}^{N_\mathcal{Z}}p\left(z_i,z_j\right) 
    \label{eq:mrf:energy}
\end{split}
\end{equation}
The first unary term penalizes discrepancy between distance from the camera to the node estimated with monocular depth CNN ($\Delta_{ij}$) and with triangulation ($d_{ij}$): 
$u_1(z_i)=z_i \sum_{j=1,2}\|\Delta_{ij}-d_{ij}\|$. 
The $u_0$ unary term promotes inclusion of terms on any ray $R$ and is inversely proportional to the number of intersections on the ray:
$u_0(z_i)= -z_i/(\sum_{x_n \in R}1)$. 
The presence of this term ensures the minimum of energy is not reached on empty configuration $z_i=0$ for all $i$.
The pairwise term penalizes distances between any two nodes simultaneously active on any view-ray $R$:
$p(z_m, z_n)= \sum_{x_m, x_n \in R} z_m z_n \|x_m-x_n\|,$ where $x_i$ is a GPS position of node controlled by state $z_i$. 
The novel $u_2$ unary term is exploiting the vertical dimension by ensuring consistency between views along this dimension. It penalizes discrepancy between height estimates based on the intersecting views:
$u_2(z_i)=z_i \|h_{i1}-h_{i2}\|$.
The weights for the terms $u_0, u_1, u_2$ and $p$ are $\alpha,\beta, \lambda \in\lbrack0,1)$, and
$1-\alpha-\beta-\lambda\geq 0$, respectively.

The state $\mathbf{z}=\lbrace z_i \rbrace_{i=1,\cdots,N_\mathcal{Z}}$ that minimizes the energy $\mathcal{U}$ (Eq. \ref{eq:mrf:energy}) is computed using the Quadratic pseudo-Boolean optimization (QPBO) algorithm~\cite{rother2007optimizing}, which is based on graph-cut optimizations. QPBO has two substantial advantages over the original Simulated Annealing based optimization proposed in~\cite{Krylov_2018}: firstly, the optimization result is deterministic, which avoids the need to rerun the procedure multiple times to arrive at an averaged (stable) field configuration. Secondly, the procedure has predictable complexity and is fast. The use of QPBO is possible by eliminating higher-order penalty terms originally used in~\cite{Krylov_2018} that did not allow for energy function composition compatible with this type of optimization. Specifically, the  stand-alone penalty
term $u_3$ used in~\cite{Krylov_2018} has been replaced by the unary term $u_0$ (Eq. \ref{eq:mrf:energy}). The pairwise term is supermodular~\cite{rother2007optimizing}, but the number of such terms is relatively small, and empirically allows for optimal state assignments to all graph nodes using QPBO/QPBOP algorithm.

\subsection{Clustering}
\label{sec:clustering}
A separate MRF graph is constructed for each object class segmented in images. This reduces the computation complexity and significantly mitigates the impact of noise when objects of different types are clustered near the same location. Specifically, due to the uncertainty/errors in the image metadata and the segmentation maps, one object can be represented by multiple positive nodes geolocated in the same area, if this object has been discovered in 3 or more images. For this reason, hierarchical clustering is employed to group together positive node locations in the same vicinity, and each object instance is represented by its cluster center. 
The downside of clustering is that it can reduce multiple real objects of the same class that are in close proximity to a single instance prediction. We have addressed this latter problem by counting the number of nodes that originated from the same pair of images in each cluster, and then splitting every cluster into a number of subclusters given by the highest count of pairs inside it.

\section{Inference of Height}
\label{sec:height}

\begin{figure}[!h]
  \centering
  \includegraphics[width=.75\linewidth]{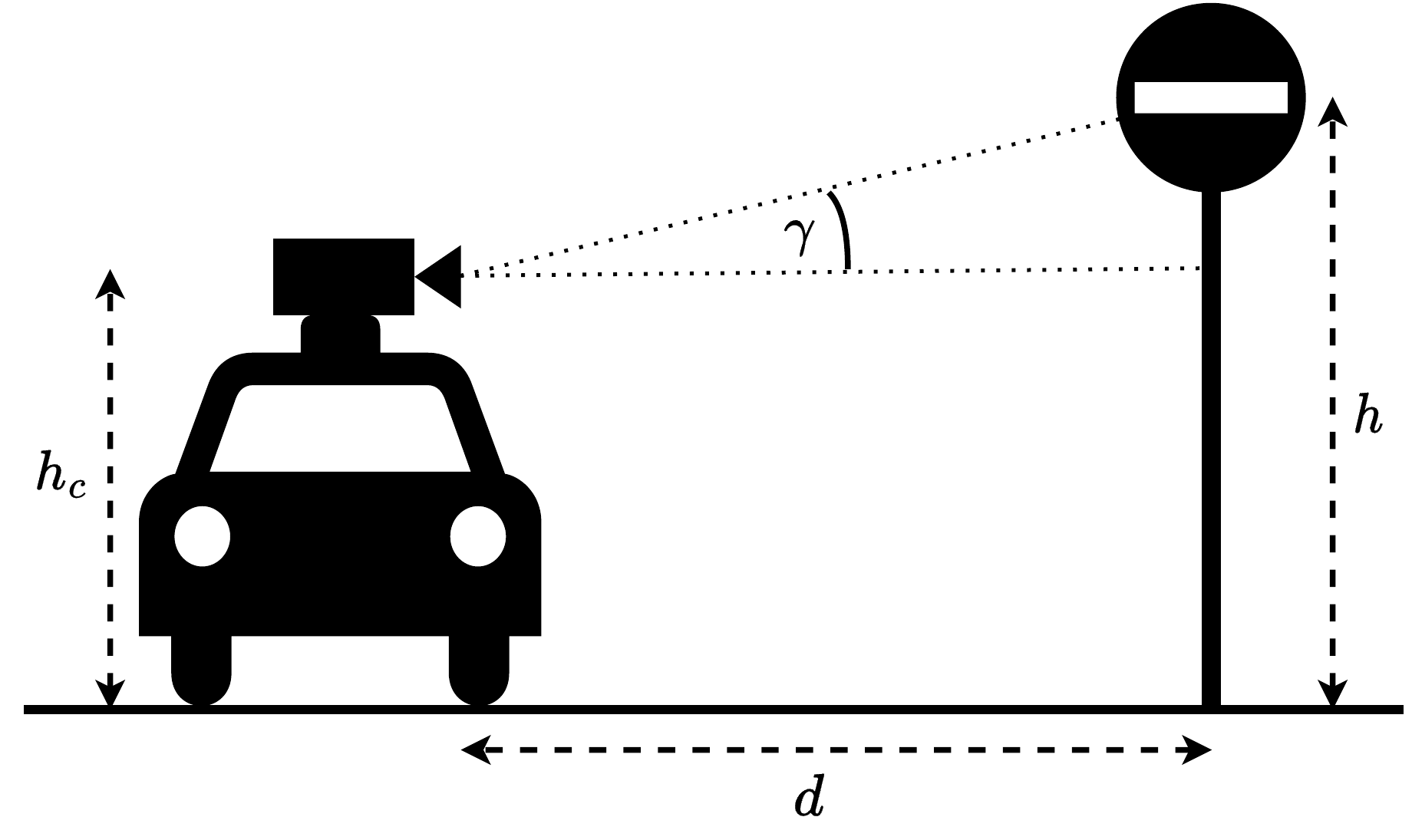}
  \caption{Object height estimation schematics.}
  \label{fig:eq:h}
\end{figure}

\begin{figure}[!h]
\centering
  \includegraphics[width=.75\linewidth,trim={0 3.5cm 0  2cm},clip]{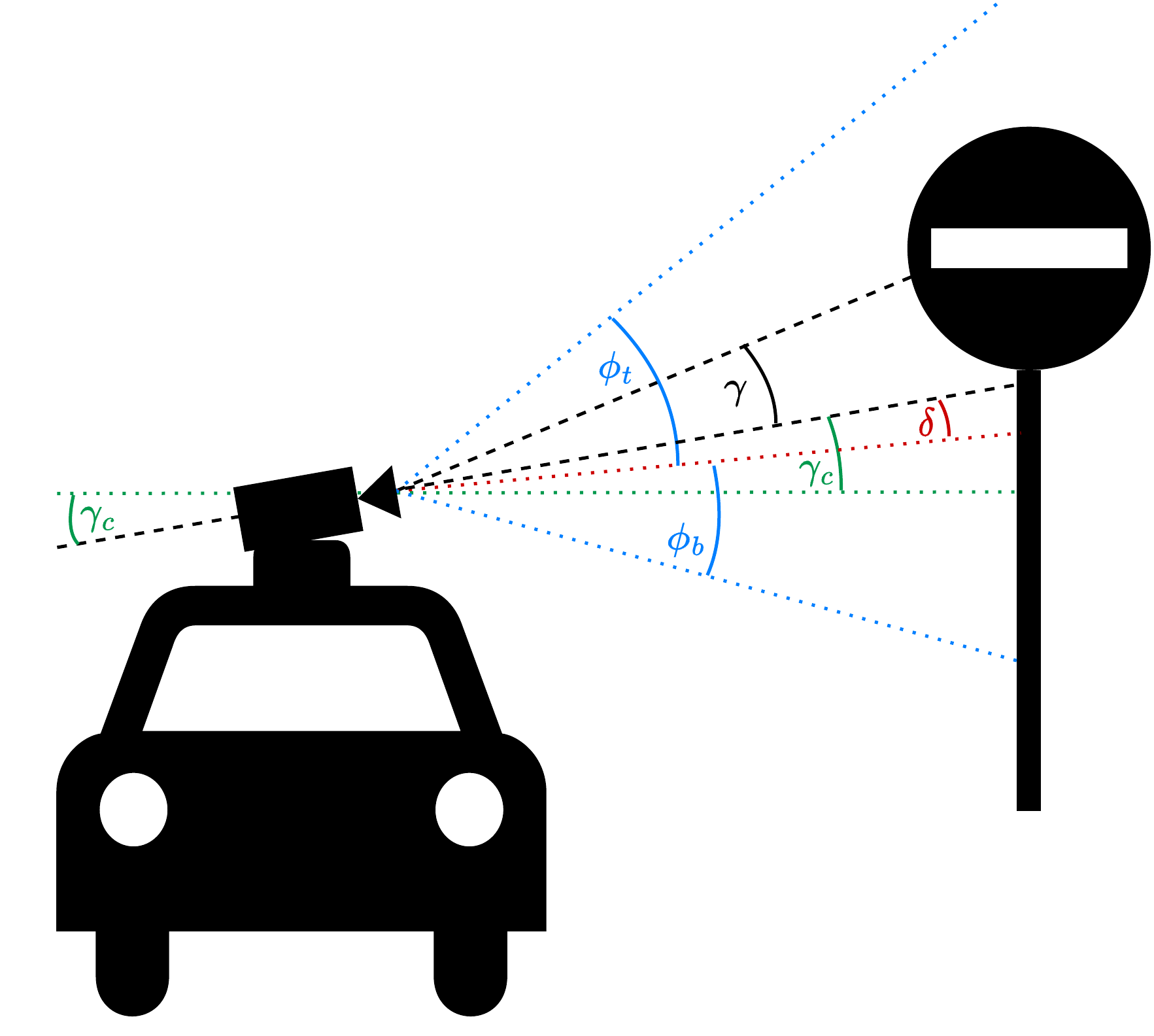}
  \caption{Angle terms for pitch estimation defined by Equation \ref{eq:correction_formula}.}
  \label{fig:eq:3}
\end{figure}

Once the object geolocation has been computed by the triangulation pipeline, its elevation can be estimated from each camera view in which the object was detected as follows:
\begin{equation}
  h = d \tan{\gamma} + h_c
  \label{eq:h}
\end{equation}
where $h_c$ is the camera elevation offset, $d$ is distance to object from the camera, and $\gamma$ is object pitch angle. Since the GPS coordinates of the camera are captured and objects of interest are geolocated with our pipeline, the \textit{camera-to-object} distance $d$  can be computed (cf. Fig. \ref{fig:eq:h}) and, provided the camera height $h_c$ and angle $\gamma$ are known, height $h$ can be computed with Eq. (\ref{eq:h}).
We have used the Haversine formula to calculate distance between two points given by their latitude and longitude. 
 To obtain pitch $\gamma$, Eq.~(\ref{eq:pitch_formula})~and~(\ref{eq:correction_formula}) are proposed here and both are tested experimentally (cf. Sec. \ref{sec:experiments}). The  CNN for object detection in an image  gives us the pixel coordinates of  the object of interest $(x, y)$ (measured to the center of the segmented object from top-left image corner). Coordinates from a rectilinear image with height $H$ are translated into pitch angle $\gamma$ employing the following formula:
\begin{equation}
  \gamma = \left \lbrack\frac{\frac{H}{2} - y}{\frac{H}{2}}\right\rbrack \frac{\phi}{2} - \gamma_c.
  \label{eq:pitch_formula}
\end{equation}
The angle $\gamma$ can be estimated when camera field of view $\phi$ and its pitch $\gamma_c$ are known (either provided, or estimated via calibration). 
Eq.~(\ref{eq:pitch_formula}) assumes homogeneous lens setup, however, in practice the camera center might not correspond to the center of the image. It is possible to devise a correction mechanism by incorporating difference between $\phi_t$ top and $\phi_b$ bottom field of view into the object pitch (see Fig. \ref{fig:eq:3}). 
Thus, Eq.~(\ref{eq:pitch_formula}) can be corrected as follows:
\begin{equation}
  \gamma = \left \lbrack\frac{\frac{H}{2} - y}{\frac{H}{2}} \right\rbrack \frac{\phi}{2} - \gamma_c + \underbrace{\frac{\phi_t - \phi_b}{2}}_{\delta}.
  \label{eq:correction_formula}
\end{equation}
An object geolocated in our pipeline that was detected $n$ times in $n$ images, will then be associated with $n$ estimates of heights in our system. The final height estimate for an object is the mean or median of the height measurements.

\section{Experimental results}
\label{sec:experiments}

\subsection{Data \& methodology}
\label{sec:data}
 
For experimental assessment we employ street level imagery with metadata and LIDAR scans that were captured in Dublin (Ireland) in  2020 by Murphy Geospatial~(\url{https://murphygs.com}) on Guild and Macken streets, each driven in both directions. The two sequences have 363 and 350 capture points, sampled approx. every 3 meters. Each sample location has 6 views captured with field of view of 68.77 degrees, covering the entire panorama. Each image has a pair of GPS coordinates attached. Image resolution is 2046$\times$2046 pixels. LIDAR data has been collected simultaneously with the optical imagery acquisition. Images and associated metadata are processed with our pipeline, while LIDAR data has been used exclusively for manually creating ground truth for object heights.
The proposed height prediction pipeline is evaluated on two types of objects, namely water drains and traffic signs. Contribution of the object height to its geolocation (i.e. the use of the proposed unary term $u_2$ in Eq.~\ref{eq:mrf:energy}) is also investigated. 

\begin{figure}[!h]
  \centering
  \subfloat{\includegraphics[width=.46\linewidth,trim={0 0 0 0.3cm},clip]{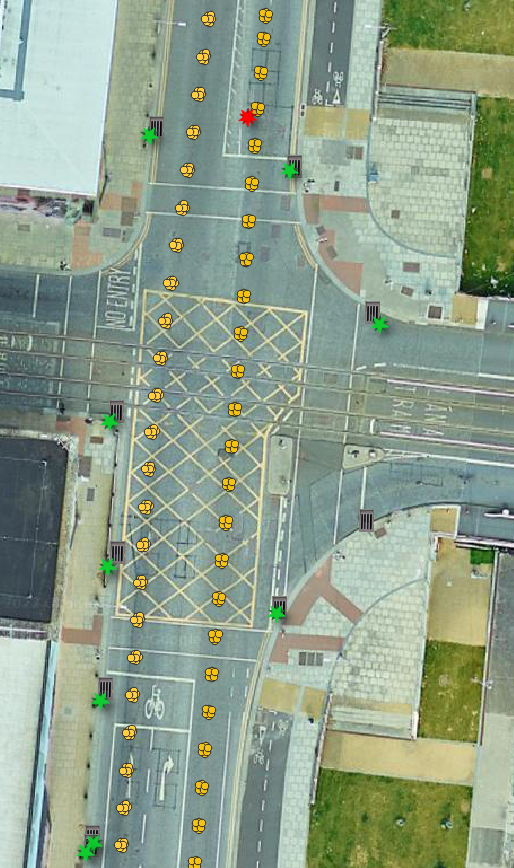}}
  \hfil
  \subfloat{\includegraphics[width=.46\linewidth]{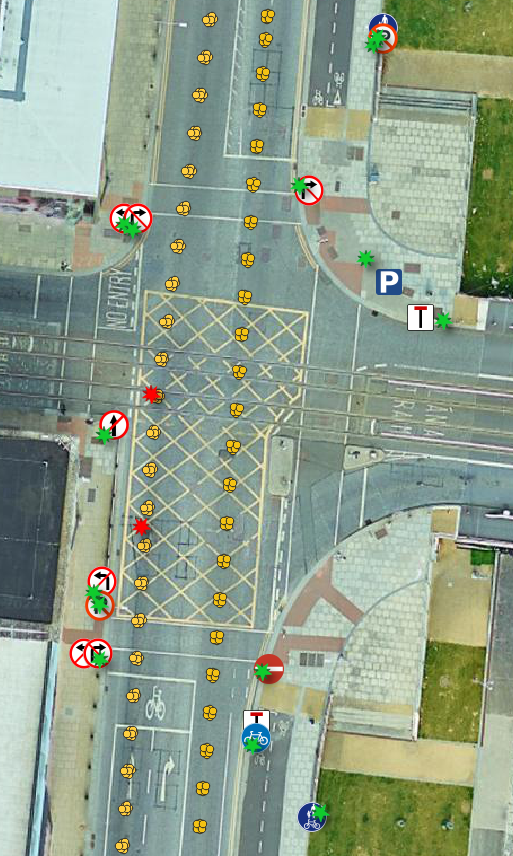}}
  \caption{QGIS aerial view with detections overlaid: water drains (left), traffic signs (right); ground truth positions (miniatures), correct detections (green stars), false positives (red stars) and car locations (orange dots).}
  \label{fig:detects_map}
\end{figure}

\subsection{Water drain detection}
\label{sec:drain}

\subsubsection{Experimental setup} The water drain detection CNN is trained on a dataset of water drains collected in Dublin city by our team, using mobile phones, captured outside of the test area. This training dataset consists of 880 images.
To create ground truth, we have assumed elevation of 0 meters (ground level) for the drains and we manually geotagged a set of 134 visible drains in the test area using satellite imagery.
Only detections within 30 meters from camera are used to generate nodes in the MRF model.
To assess geolocation, the data is split (by latitude separation) into validation and test set with 66 and 68 drains respectively. Hyperparameters $\alpha, \beta, \lambda$ and cluster size limit are found via grid search and configuration that maximizes f-score on the validation set is selected.
The estimated elevation is assessed (using optimal hyperparameters for the entire dataset) only for the true positive predictions, which are defined as those within 6 meters from ground truth. 
\begin{table*}[!h]
\centering
\tabcolsep = 1.7mm
\caption{Geolocation results when considering monocular depth term $u_1$, and height term $u_2$ in MRF energy (other terms $u_0$ and $p$ are included in all configurations). Positional errors are reported in meters.}
\label{tab:geolocalization}
\vspace{-.7em}
\begin{tabular}{l|c|c|c|c|c|c|c|c|c|c|c}
 \hline
 \multirow{2}{*}{MRF terms} & \multirow{2}{*}{object} & \multicolumn{5}{c|}{validation} & \multicolumn{5}{c}{test} \\
   &  & precision$\uparrow$ & recall$\uparrow$ & f-score$\uparrow$ & error mean$\downarrow$ & error median$\downarrow$ & precision$\uparrow$  & recall$\uparrow$ & f-score$\uparrow$ & error mean$\downarrow$ & error median$\downarrow$ \\
 \hline
 $u_1$ (depth) & drain & 0.984 & 0.909 & 0.945 & 0.832 & 0.712 & 0.956 & 0.956 & 0.956 & 0.857 & 0.804 \\
 $u_2$ (height) & drain & 0.917 & 0.833 & 0.873 & 1.483 & 0.979 & 0.800 & 0.824 & 0.812 & 1.415 & 1.003 \\
 $u_1 + u_2$ & drain & 0.984 & 0.939 & \bf{0.961} & 0.945 & 0.829 & 0.932 & 1.000 & \bf{0.965} & 0.920 & 0.862 \\
 \hline
 $u_1$ (depth) & sign & 0.738 & 0.842 & 0.787 & 1.551 & 1.007 & 0.850 & 0.895 & \bf{0.872} & 1.171 & 0.993 \\
 $u_2$ (height) & sign & 0.628 & 0.860 & 0.726 & 1.459 & 1.087 & 0.769 & 0.877 & 0.820 & 1.227 & 0.958 \\
 $u_1 + u_2$ & sign & 0.735 & 0.877 & \bf{0.800} & 1.526 & 1.060 & 0.836 & 0.895 & 0.864 & 1.094 & 1.013 \\
 \hline
\end{tabular}
\end{table*}

\subsubsection{Results}
As demonstrated in Table~\ref{tab:geolocalization}, height information ($u_2$ term) does not provide a meaningful replacement for monocular depth information ($u_1$), but can provide relevant complementary information that improves the localization performance.
Table~\ref{tab:drains} shows that mean estimated elevation across all of the true positive drains at ground level with respect to the camera is ~-2.18 meters. This information gives us an estimate of the height $h_c$ of the camera that can be used subsequently to infer heights of objects above ground level. Correcting vertical angle of detected objects (pitch) with the pitch of the camera reduces the average standard deviation of elevation predictions of an object from different viewpoints. The bottom row of Table~\ref{tab:drains} shows that using height information in the MRF energy leads to more stable height estimation by further reducing the standard deviation.
\begin{table}[!t]
\centering
\caption{Water drain elevations (in meters)  computed  using Eq. (\ref{eq:h}) with different computation of  $\gamma$. When $\gamma_c\neq 0$ and $\delta\neq 0$, these parameters are found using the image metadata.}
\label{tab:drains}
\vspace{-.7em}
\begin{tabular}{l|c|c|c}
 \hline
 pitch correction & median elev. & mean elev. & std  \\
 \hline
 Eq. (\ref{eq:pitch_formula})   $\gamma_c=0$ & -2.177 &  -2.179 & 0.159 \\
 Eq. (\ref{eq:pitch_formula})  $\gamma_c\neq 0$ & -2.157 & -2.167 & 0.134 \\
 Eq. (\ref{eq:correction_formula}) $\gamma_c,\delta\neq 0$ & -2.103 & -2.121 & 0.123 \\
 Eq. (\ref{eq:correction_formula}) $\gamma_c,\delta\neq 0$,$u_2$ & -2.095 & -2.103 & 0.089 \\
 \hline
\end{tabular}
\end{table}

\subsection{Traffic sign detection}
\label{sec:signs}

\subsubsection{Experimental settings}
To perform traffic sign detection we train the CNN model on the MTSD dataset~\cite{ertler2020mapillary}. We omit the complementary traffic sign classes and aggregate the rest into 174 classes according to their meaning.
Similar to drains, a ground truth set of 114 traffic signs was annotated with their particular classes. 
We manually collect ground truth heights from the point cloud created by the LIDAR scan.
The elevation is measured as the height of the sign from the ground to the top of the sign less half of the sign's vertical size. 
The data is split again to validation and test set, each consisting of 57 signs.
A detected traffic sign is considered to be true positive only if its class is predicted correctly. In this experiment, the elevation $h_c$ used for  the camera height is the one found previously as a result of the experiment on water drains, i.e. 2.18 meters.

\subsubsection{Results}
Geolocation results in Table~\ref{tab:geolocalization} align with the findings on the drain object detection.
The precision and recall are lower than those reported for the drains, since the used detector was trained on signs collected globally, and only a portion of the training samples correspond to the traffic signs in use in Ireland.
Fig.~\ref{fig:detects_map} visualizes the detections on a section of the surveyed street.
Table~\ref{tab:signs} lists height calculation results using median and mean estimation based on the correctly detected signs. Correcting the tilt and using vertical information in the localization process ($u_2$) reduces not only the average standard deviation as in case of drains, but also average prediction error. The average estimation error of sign elevation is approximately 20cm.
\begin{table}[!h]
\centering
\caption{Traffic sign height estimation results (in meters).}
\label{tab:signs}
\vspace{-.7em}
\begin{tabular}{l|c|c|c}
 \hline
 pitch correction & med. pr. err. & mean pr. err. & std \\
 \hline
 Eq. (\ref{eq:pitch_formula})   $\gamma_c=0$  & 0.238 & 0.279 & 0.255 \\
 Eq. (\ref{eq:pitch_formula})  $\gamma_c\neq 0$  & 0.192 & 0.220 & 0.178 \\
 Eq. (\ref{eq:correction_formula}) $\gamma_c,\delta\neq 0$ & 0.171 & 0.201 & 0.162 \\
 Eq. (\ref{eq:correction_formula}) $\gamma_c,\delta\neq 0$, $u_2$ & 0.151 & 0.195 & 0.150 \\
 \hline
\end{tabular}
\end{table}
Note that errors in distance estimation originating from potential triangulation inaccuracies directly affect estimated elevation. We pick the center of the object for ground truth elevation, however, errors obtained are on average smaller than the size of the inspected objects and thus the estimated elevation still often falls into the extent of the object. Our results are a slight improvement over Ning et al.~\cite{Ning_2022} that reported 0.218 meter average error when estimating above-ground  elevation of house doors using GSV imagery. 

\section{Conclusion}
We have extended the object localization pipeline to predict elevation of street objects and furniture from street level images at negligible increase in computation cost, by reusing  image metadata. The new formulation incorporates heights information into the geolocation process and has been extended to a multi-class object detection scenario. We have achieved average height estimation error lower than 20cm, as tested on traffic signs and water drains in urban environment.

\bibliographystyle{IEEEbib}
\bibliography{bibliography}

\end{document}